\documentclass[runningheads]{llncs}
\usepackage{graphicx}

\usepackage{tikz}
\usepackage{comment} 
\usepackage{amsmath,amssymb} 
\usepackage{color}

\usepackage{subfigure}
\usepackage{booktabs}
\usepackage{multirow}
\usepackage{pifont} 
\newcommand{\cmark}{{\color{black}\ding{51}}}%

\usepackage[pagebackref=true,breaklinks=true,letterpaper=true,colorlinks,bookmarks=false]{hyperref}

\def\etal{\emph{et al.}~}

\begin{document}
\pagestyle{headings}
\mainmatter
\def\ECCVSubNumber{1078}  

\title{Object-and-Action Aware Model for Visual Language Navigation} 

\titlerunning{OAAM VLN}
%
\author{Yuankai Qi\inst{1} \and
Zizheng Pan\inst{1} \and
Shengping Zhang\inst{2,3} \and
\\
Anton van den Hengel\inst{1} \and
Qi Wu$^*$\inst{1} 
}
\authorrunning{Y. Qi et al.}
%
\institute{The University of Adelaide\\
\and
Harbin Institute of Technology, Weihai \and
Aipixel Inc.\\
\email{qykshr@gmail.com, zizhpan@gmail.com, s.zhang@hit.edu.cn\\
	\{anton.vandenhengel, qi.wu01\}@adelaide.edu.au}}
\maketitle

\begin{abstract}
Vision-and-Language Navigation (VLN) is unique in that it requires turning relatively general natural-language instructions into robot agent actions, on the basis of visible environments.   
This requires to extract value from two very different types of natural-language information. 
The first is object description (e.g., `table', `door'), each presenting as a tip for the agent to determine the next action by finding the item visible in the environment,
and the second is action specification (e.g., `go straight', `turn left') which allows the robot to directly predict the next movements without relying on visual perceptions. 
However, most existing methods pay few attention to  distinguish these information from each other during instruction encoding and mix together the matching between textual object/action encoding and visual perception/orientation features of candidate viewpoints.
In this paper, we propose an Object-and-Action Aware Model (OAAM) that processes these two different forms of natural language based instruction separately.  
This enables each process to match object-centered/action-centered instruction to their own counterpart visual perception/action orientation flexibly. 
However, one side-issue caused by above solution is that an object mentioned in instructions may be observed in the direction of two or more candidate viewpoints, thus the OAAM may not predict the viewpoint on the shortest path as the next action. To handle this problem,  we design a simple but effective path loss to penalize trajectories deviating from the ground truth path. 
Experimental results demonstrate the effectiveness of the proposed model and path loss, and the superiority of their combination  with a $50\%$ SPL score on the R2R dataset and a $40\%$ CLS score on the R4R dataset in unseen environments, outperforming the previous state-of-the-art.
\let\thefootnote\relax\footnotetext{$^*$corresponding author}
\keywords{Vision-and-Language Navigation, Modular Network, Reward Shaping}
\end{abstract}

\section{Introduction}
Vision-and-language navigation (VLN) has attracted increasing attention, partly due to the prospect that the capability to interpret general spoken instructions might represent a significant step toward enabling intelligent robots~\cite{r2r,sf,trlvln,regret}.  
The goal of VLN is for a robot to interpret a navigation instruction expressed in natural language, and carry out the associated actions.
Typically this involves an agent navigating through a 3D photo-realistic  simulator~\cite{r2r} to a goal location, which is not visible from the start point.  The natural language instructions involved are of the form of `Go left down the hallway toward the exit sign until the end. Turn right and go down the hallway. Go into the door on the left and stop by the table.'.  The simulator in this scenario provides a set of panoramic images of a real environment, and a limited set of navigation choices that can be taken for each.
Although it is related to other Vision-and-Language problems that have been extensively studied, VLN remains an open problem due to the 
difficulty of navigating general real environments, and the complexity of the instructions that humans give for doing so.

\begin{figure}[!t]
	\centering  
	\includegraphics[width=1.0\linewidth]{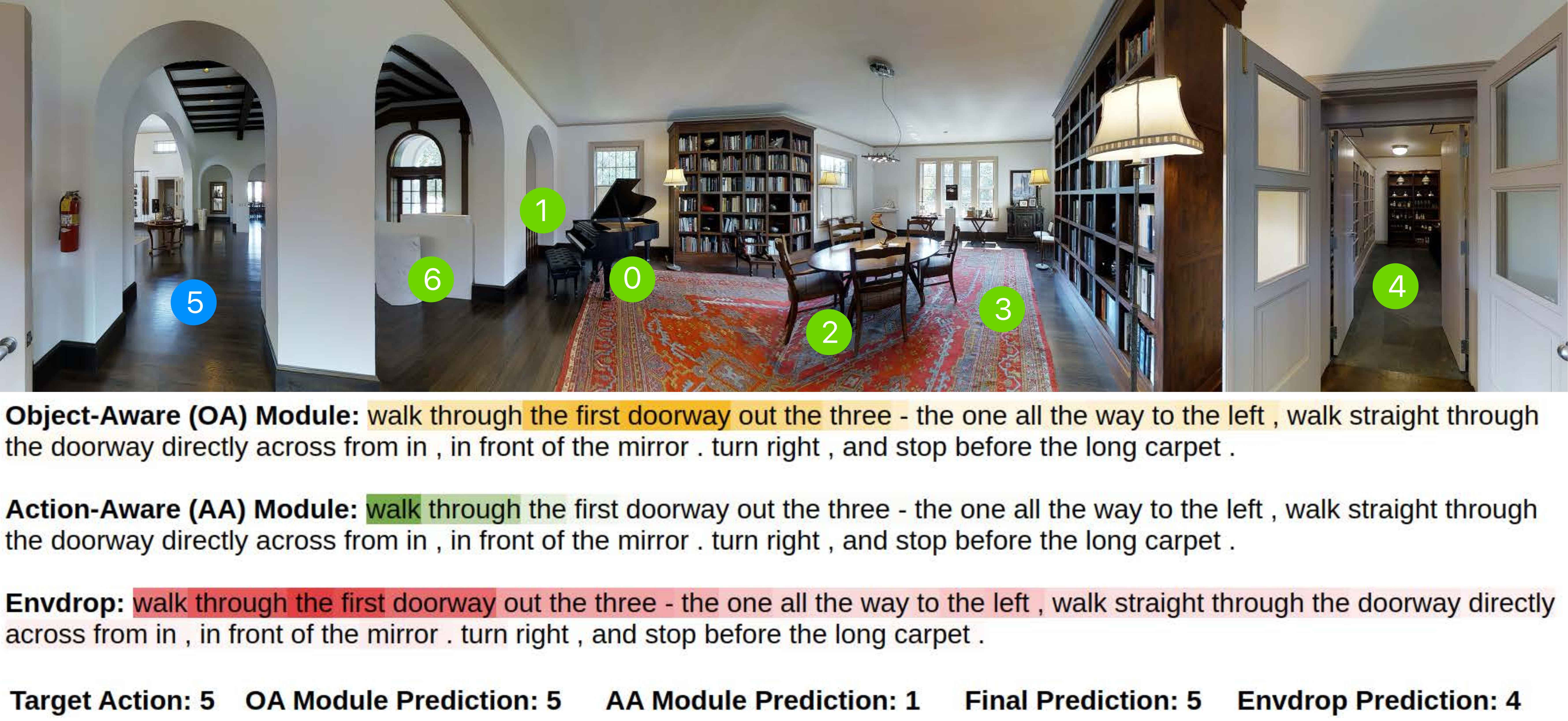}
	\caption{A snapshot of navigating action predicted by the proposed  model and  our baseline navigator EnvDrop~\cite{envdrop}. Our model is able to more flexibly utilize object  and action phrases thanks to the object-/action-  aware modules.  {The numbered circles} denote navigable viewpoints, and the blue one is our final decision.
	}
	\label{fig:firstShow}
\end{figure}

A range of approaches have been proposed to address the vision-and-language guided navigation  problem~\cite{selfMonitor,sf,envdrop,regret,emnlp,lbyl,wx2,wx3}. For instance, to progressively process long navigation instructions, Ma \etal \cite{selfMonitor} propose a progress monitor network to identify the completed part of an instruction and the part associated with the next action. To expand training data, the Speaker-Follower (SF) \cite{sf} and EnvDrop \cite{envdrop} are proposed to generate new trajectories and `unseen' scenes from seen ones, respectively. To mimic human behaviors when navigating, the Regretful model \cite{regret} introduces a backtracking mechanism into the navigation process to enable the agent to retrace its steps. In FAST model~\cite{fast}, a strategy that compromises between greedy search and beam search is developed to balance search efficiency against accuracy. 
However, all of these approaches entangle the encoding of object descriptions with that of action specifications, instead of processing them separately.
Most action specifications (e.g., `go straight ahead' or `turn right'), closely relating to orientations$^1${\let\thefootnote\relax\footnotetext{$^1$`orientation' means the encoding of the angle that an agent should rotate in order to find a navigable viewpoint from its front direction as in~\cite{fast,selfMonitor,regret}.}}  rather than visual perceptions of each navigable candidate viewpoint, are able to directly lead to the correct next action. 
By contrast, the object descriptions (e.g., `the stairs' or `the table') correspond to visual perceptions, instead of the orientations, in the visible scene. 
Therefore, the mixed encoding scheme may limit the similarity/grounding learning in the decoding phase between instructions and visual perceptions as well as orientations.

To address the above mentioned problem,
we propose here an object-and-action aware model (OAAM) for robust navigation that reflects to exploit the important distinction. 
Specifically, we first utilize two learnable attention modules to highlight language relating to objects and actions within a given instruction. Then, the resulting attention maps  guide an object-vision matching module and an action-orientation matching module, respectively.  
Finally, the action to be taken is identified by combining the outcomes of both modules using weights derived by an additional attention module taking the instruction and visual context as input.
Figure~\ref{fig:firstShow} provides an example of the VLN process that demonstrates that our design enables the agent to more fleixibly utilize object descriptions and action specifications, and finally leads to the correct prediction compared against the strong baseline navigator EnvDrop~\cite{envdrop}.

The proposed OAAM is able to improve the navigation success as demonstrated later in the experiment. However, its trajectory might not be the shortest because  the focused object in the instruction may be observed in the direction of multiple candidate viewpoints. 
To handle this problem, we design a simple but effective path loss to encourage the agent to stay on the shortest path instead of merely picking alternative viewpoints containing the instruction mentioned object. 
In particular, the proposed path loss  is based on  distances calculated at each agent step to its nearest viewpoint on the ground-truth path. Note that this differs from the Coverage weighted by Length Score (CLS) award~\cite{r4r} that computes a normalised score by inversely finding the nearest viewpoint in the agent trajectory to each viewpoint on the ground-truth path.
Experimental results show that this loss aids the OAAM to generalise in unseen scenarios.

The main contributions of this work are summarized as follows:
\begin{enumerate}
	\setlength{\itemsep}{1.5ex}  
	\setlength{\itemindent}{1.1em}  
	\item We propose an object-and-action aware model, which better reflects the nature of the associated natural language instructions and responds  more flexibly.
	\item We design a path loss that encourages the agent to closely follow navigation instructions.
	\item Extensive experimental results demonstrate the effectiveness of the proposed method and set new state-of-the-art on the R4R dataset with a CLS score of $0.40$.
\end{enumerate}

\section{Related Work}

\noindent\textbf{Vision-and-Language Navigation.} 
Numerous methods have been proposed to address the VLN problem. Most of them employ the CNN-LSTM architecture with attention mechanism to first encode instructions and then decode the embedding to a series of actions.
Together with the proposing of the VLN task,
Anderson \etal~\cite{r2r} develop the teacher-forcing and student-forcing training strategy. The former equips the network with basic ability for navigating using ground truth action at each step, while the latter mimics the test scenarios, where the agent may predict wrong actions, by sampling actions instead of using ground truth actions. 
Deep learning based methods always benefit from massive training data. To generate more training data, Fried \etal \cite{sf} propose a speaker model to synthesize new instructions for randomly selected robot trajectories.
By contrast, Tan \etal ~\cite{envdrop} augment the training data by additionaly removing objects from the original training data to generate new `unseen' environments. 
To further enhance the generalization ability, they also propose to train the model using both imitation learning and reinforcement learning so as to take advantage of both off-policy and on-policy optimization.
Not all chunks of an instruction are useful to predict the next action. Ma \etal \cite{selfMonitor} propose a progress monitor to locate the completed sub-instructions and to be aware of the ones for predicting the next action.
Another way to improve navigation success is to equip the agent with backtracking.
In \cite{regret}, Ma \etal propose to treat the navigation as a graph search problem and predict to move forward or roll back to a previous viewpoint. In \cite{fast}, each passed viewpoint is viewed as a goal candidate, and the final decision is the viewpoint with the highest local and global matching score.  

However, the above mentioned methods neglect to distinguish the object descriptions and the action specifications within the focused sub-instruction. This may limit the learning of matching between object-/action-centered instructions and their counterpart visual perceptions/orientations.
To address this problem, we propose to separately learn the object attention and action attention, and so the learning of object-vision matching and action-orientation matching. 

\begin{figure*}[!t]
	\centering
	\includegraphics[width=\textwidth]{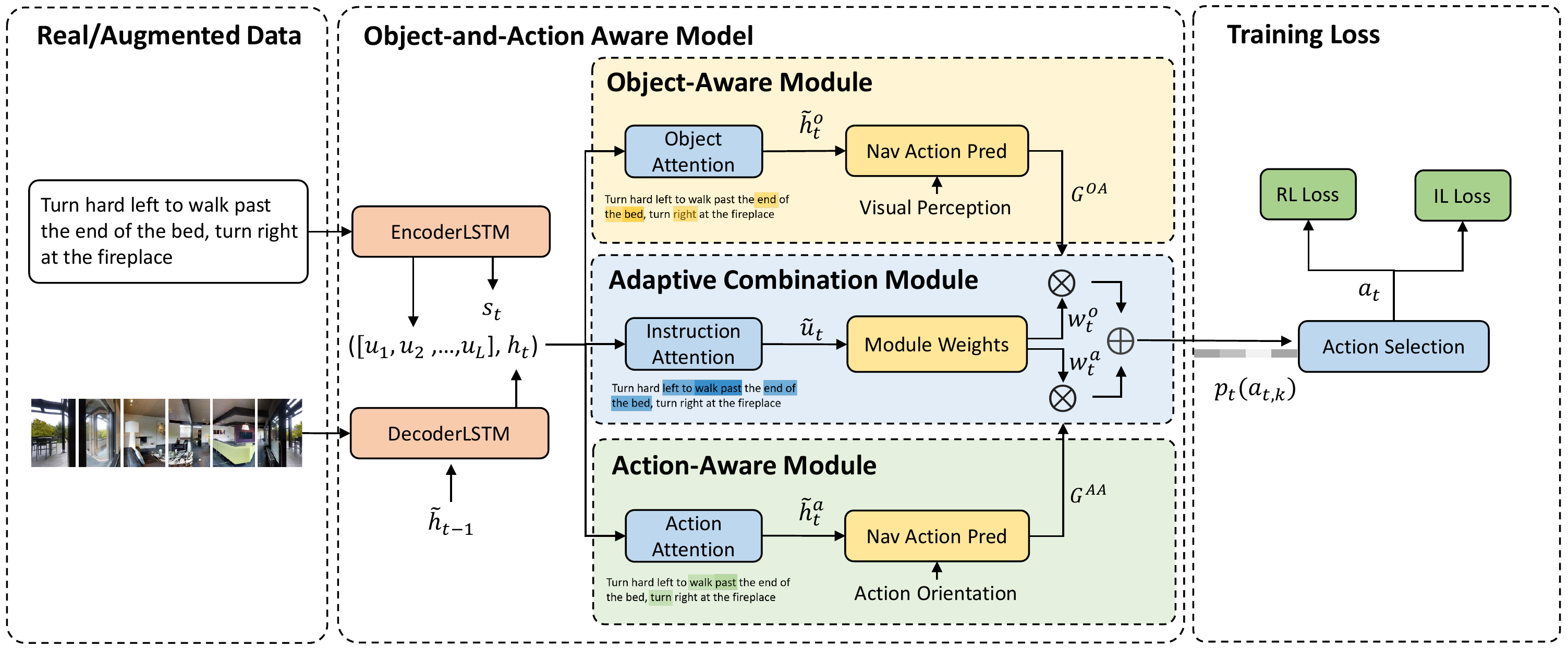}
	\caption{Main steps of the proposed object-and-action aware model, which consists of three modules: the object-aware module that predicts the next action based on object-centered instructions and visual perceptions; the action-aware module that predicts the next action based on action-centered instructions and orientations of candidate viewpoints; and the adaptive combination module that predicts weights to combine predictions obtained by the other two modules.
	}
	\label{fig:model_info}
\end{figure*}

\noindent\textbf{Modular Language Process Networks.} 
Modular language  decomposition  has been widely adopted in question answering~\cite{mqa}, visual reasoning~\cite{mvqa}, referring expression comprehension~\cite{mrm,mattnet}, etc. Most methods are based on hard parsers or learned parsers. For example, Andreas \etal ~\cite{mqa}  decompose each question into a sequence of modular sub-problems with the help of an fixed off-the-shelf syntactic parsers. Such hard parsers are not able to analyse semantic relationships between entities. By contrast, the learned parses can be flexibly designed according to parsing purposes. For instance, Hu \etal \cite{mrm} design a attention-based network to learn to decompose expressions into (Subject, Relationship, Object) triplets, which facilitates to ground language to visual regions. In \cite{mattnet}, Yu \etal propose the (Subject, Location, Relationship) triplet to additionally taking object position into consideration.

Inspired by the above modular decomposition mechanism, we propose to decompose an instruction into object and action components for the VLN task because two fundamental different type of information of candidate viewpoints are used to match to instructions, namely visual perceptions and orientations. Differ from the above mentioned methods, our modular decomposition model takes navigation progress into consideration, i.e., the decomposition should take place on a certain part of the instruction, which is closely related to the current agent location, instead of the whole instruction.

\section{Method}

In this section, we first describe the symbols that will be used later and briefly introduce the baseline navigator, EnvDrop~\cite{envdrop},  to put our method into a proper context.   EnvDrop is adopted   due to its good performance. Then, we detail the proposed object-and-action aware model (OAAM).

Figure~\ref{fig:model_info} shows the pipeline of the proposed method in training phase. An instruction is first encoded by a bi-direction LSTM. Then our OAAM model decomposes the instruction encoding into object- and action-centered representations using two attention-based modules. These representations are further used to predict navigation actions via matching to visual percenptions and orientations, respectively. The OAAM model also predicts combination weights to compute the final navigation action probability distribution over candidate viewpoints. Lastly, imitation learning and reinforcement learning losses are calculated based on the probability distribution and a proposed path loss. In the inference phase, the navigation action with maximum probability is selected.

\subsection{Problem Setup}
Given   a natural language instruction $I = \{w_1, w_2, ..., w_L\}$ with $L$ words, at each step $t$, the agent   observes a panoramic view \(o_t\), which contains 36 discrete single views \(\{o_t\}_{i=1}^{36}\). Each view \(o_{t,i}\) is represented by an image \( v_{t,i} \) at an orientation with heading angle \( \theta_{t,i}\) and   elevation angle \( \phi_{t,i}\). At each step $t$, there are $N_t$ navigable viewpoints $\{\textrm{P}_{t,k}\}_{k=1}^{N_t}$. The agent needs to select one viewpoint $\textrm{P}_{t,k}$ as  the next navigate action $a_t$. 
Following the common practice, each navigable viewpoint has  an orientation feature 
$
n_{t,k} = ( cos\theta_{t,k}, sin\theta_{t,k}, cos\phi_{t,k}, sin\phi_{t,k})
$
and  a visual feature
$
m_{t,k} = ResNet(v_{t,k}).
$


\subsection{Baseline--EnvDrop}
Recent works show that data augmentation is able to significantly improve the generalization ability in unseen environment~\cite{sf,envdrop}. We make use of this strategy by adopting EnvDrop~\cite{envdrop} as our baseline, which simultaneously benefits from the back translation technique~\cite{sf} to generate new (trajectory, instruction) pairs in existing environments and from its global dropout technique to generate new environments.

EnvDrop first uses a bidirectional LSTM to encode instruction $I$:
\begin{equation}
[u_1, u_2, ..., u_L] = \textrm{Bi-LSTM}(e_1, ..., e_L)
\end{equation}
where $e_j = \textrm{embedding}(w_j)$ is the embedded $j$-th word in the instruction, and \(u_j\) is the feature containing  context information for the $j$-th word.
%
Then, a soft attention is imposed on visual feature $m_t$ to get attentive feature $\tilde f_t = \Sigma_i \alpha_{t,i} m_{t,i}$, where
$\alpha_{t,i} = \textrm{softmax}_i(m_{t,i}^\top W_F \tilde h_{t-1} )$.
The concatenation of $\tilde f_t$  and the previous action embedding  $\tilde a_{t-1}$ is fed into the decoder LSTM:
\begin{equation}
h_t = \textrm{LSTM\_Decoder}([\tilde f_t; \tilde a_{t-1}],\tilde h_{t-1}).
\end{equation}
The decoder input also includes the previous instruction-aware hidden output $\tilde h_{t-1}$, which is updated based on the attentive instruction feature $\tilde{u}_t$ and the newly obtained $h_t$:
\begin{equation}
\tilde h_t = \textrm{tanh}(W_L[\tilde u_t ; h_t]),
\end{equation}
\begin{equation}
\tilde u_t = \Sigma_j \beta_{t,j} u_j,
\end{equation}
where $\beta_{t,j} = \textrm{softmax}_j(u_j^\top W_I h_t)$, and $W_L$ is trainable parameter. 
%
Finally, EnvDrop predicts navigation action by selecting a navigable view with the biggest probability
\begin{equation}
a_t^*=\mathop{\arg\max}_{k}{P(a_{t,k})},
\end{equation}
where
$
P(a_{t,k}) = \textrm{softmax}_k([m_{t,k};n_{t,k}]^\top W_B\tilde{h}_t).
$

Different from EnvDrop, our navigator calculates the probability $P(a_{t,k})$ using the proposed object-and-action aware model as detailed in the next section.

\subsection{Object-and-Action Aware Model}

To disentangle the encoding of object- and action-related instruction, as well as  the matching to their counterpart visual perceptions and orientations of candidate viewpoints, we propose the object-and-action aware model (OAAM). OAAM  consists of three key modules: an object-aware (OA) module that is aware of object-related instruction and predicts action using visual perceptions; an action-aware (AA) module that is aware of action-related instruction and predicts action using orientations of candidate viewpoints; and an adaptive combination module that learns weights to combine the action predictions from the other two modules based on the attentive instructions with the consideration of the current panoramic views.

\begin{figure*}[!t]
	\centering
	\hfill
	\subfigure[]{
		\label{fig:objactmodule}
		\includegraphics[width=0.5\linewidth]{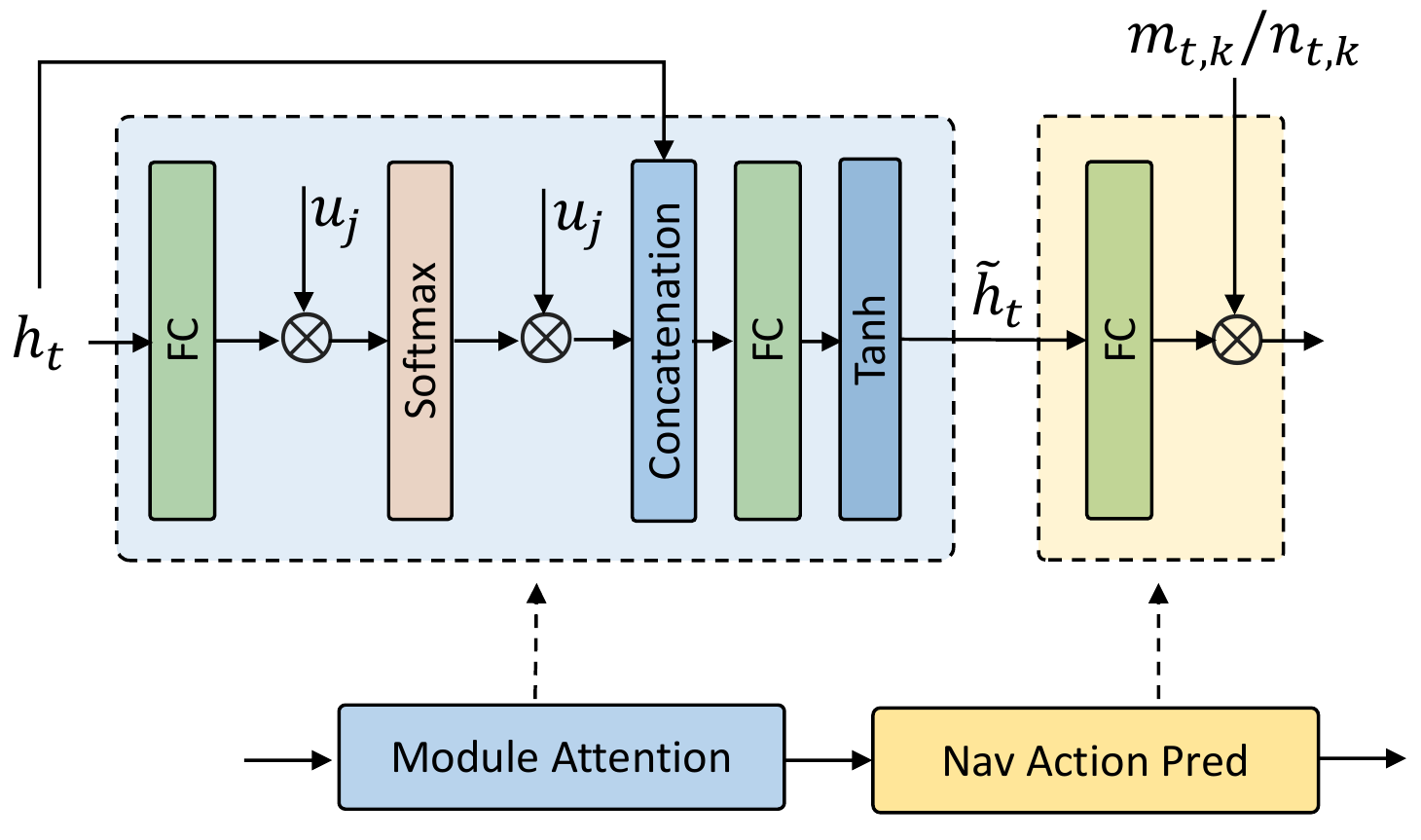}}%
	\hfill
	\subfigure[]{
		\label{fig:weightModule}
		\includegraphics[width=0.42\linewidth]{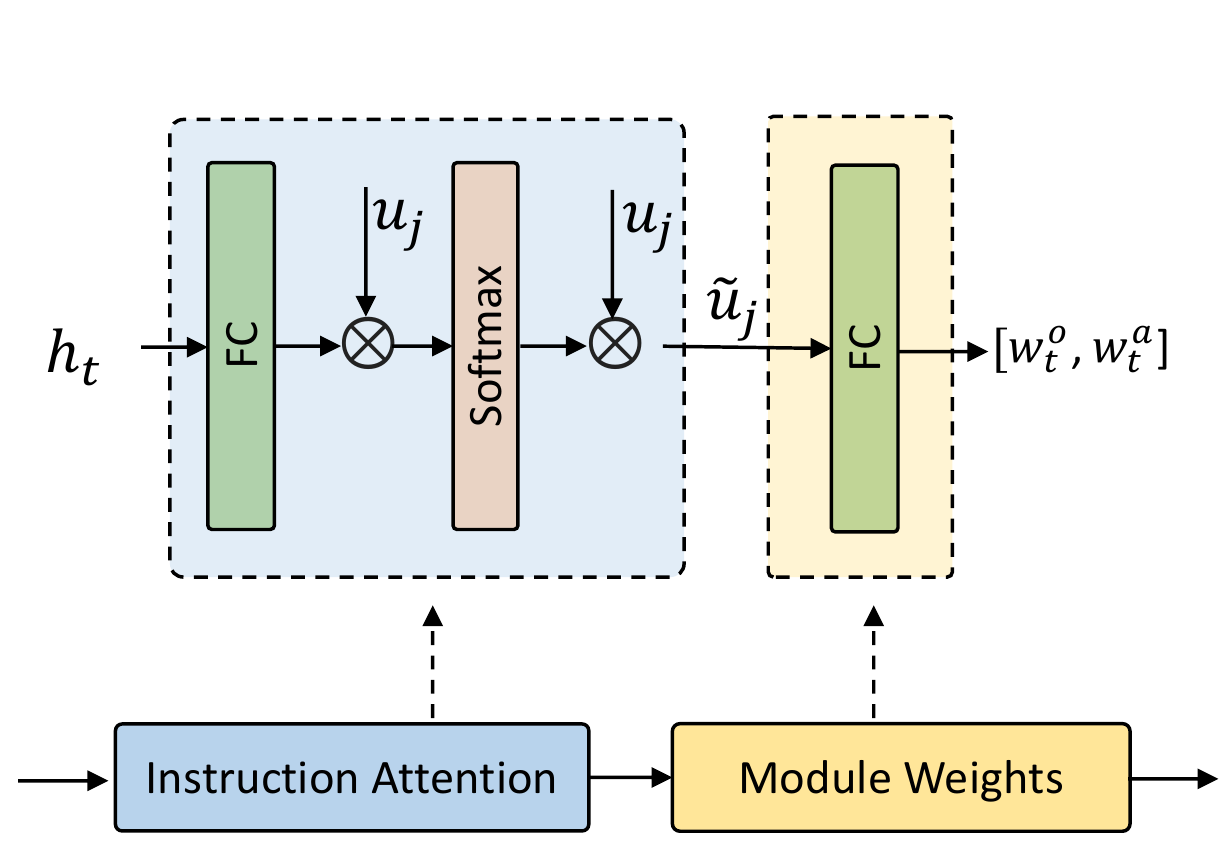}}%
	\hfill
	\caption{Pipeline and configuration of the proposed object-/action-aware module (a) and the adaptive combination module (b).
	}
	\label{fig:section33_intro}
\end{figure*}

\noindent\textbf{Object-Aware (OA) Module}
Figure~\ref{fig:objactmodule} shows the pipeline and configuration of the proposed OA module.
Different from the subject/object module in referring expression comprehension methods, such as~\cite{mre,mattnet}, which highlight all objects in an expression, our OA module is designed to highlight object phrase \textit{just for the next navigation step} instead of \textit{all objects} in an instruction. This is essential for the VLN task because the long instruction should be carried out step-by-step, which indicates objects not related to the current step may be noise and mislead the action predicting.
To this end, our OA module calculates the attentive object feature $\tilde u_t^o$ by taking the decoder hidden state $h_t$ as input for navigation progress reference:
\begin{equation}
\gamma_{t,j} = \textrm{softmax}_j(u_j^\top W_O h_t),
\end{equation}
\begin{equation}
\tilde u_t^o = \Sigma_j \gamma_{t,j} u_j,
\end{equation}
where $W_O$ is learnable parameters.
Then, the object-aware hidden output \(\tilde h_t^o\) related to the current navigation is computed via
\begin{equation}
\tilde h_t^o = \textrm{tanh}(W_P[\tilde u_t^o ; h_t]),
\end{equation}
where $W_P$ is parameter to be learned.
Lastly, the action confidence \(G^{OA}(a_{t,k})\) from the object-aware side is obtained by  using only the visual feature \(m_{t,k}\):
\begin{equation}
G^{OA}(a_{t,k}) = m_{t,k}^\top W_H \tilde h_t^o,
\end{equation}
where $W_H$ are trainable parameters.

\noindent\textbf{Action-Aware (AA) Module}
The architecture of AA module is similar to OA module, except the  visual feature \(m_{t,k}\) is replaced with the orientation feature \(n_{t,k}\) when computing the confidence of a navigable viewpoint to be the next action:
\begin{equation}
\delta_{t,j} = \textrm{softmax}_j(u_j^\top W_A h_t)
\end{equation}
\begin{equation}
\tilde u_t^a = \Sigma_j \delta_{t,j} u_j
\end{equation}
\begin{equation}
\tilde h_t^a = \textrm{tanh}(W_K[\tilde u_t^a ; h_t])
\end{equation}
\begin{equation}
G^{AA}(a_{t,k}) =n_{t,k}^\top W_F \tilde h_t^a,
\end{equation}
where \(\delta_{t,j}\) is the action attention on $j$-th word, \(\tilde h_t^a\) is action-aware hidden state, and \(W_A, W_K, W_F\) are trainable parameters. Both $\delta_{t,j}$ and $\tilde h_t^a$ have taken the navigation progress into consideration.

\noindent\textbf{Adaptive Combination Module}
Figure~\ref{fig:weightModule} shows the pipeline and configuration of the proposed adaptive combination module.
The final probability $P(a_{t,k})$ of navigable view $k$ is a softmax of weighted sum of the object-aware confidence $G^{OA}(a_{t,k})$ and action-aware probability $G^{AA}(a_{t,k})$:
\begin{equation}
\label{weightedsum}
P(a_{t,k}) = \textrm{softmax}([G^{OA}(a_{t,k}) ; G^{AA}(a_{t,k})] w_t),
\end{equation}
where $w_t$ is a predicted weight vector. $w_t$ should vary as the navigation goes because the importance of an object description and an action specification may change at different processing point of the instruction.
Thus, to adaptively combine $G^{OA}(a_{t,k})$ and $G^{AA}(a_{t,k})$ in terms of the processing state, we utilize a trainable layer to predict weights $w_t = W_T  \tilde u_t$, where \(W_T\) is trainable parameter, $\tilde{u}_t$ is the vision-aware attentive instruction feature.

\subsection{Training Loss}

\begin{figure}[!t]
	\centering  
	\hfill
	\subfigure[Distance to Target]{
		\label{goal_pathloss}
		\includegraphics[width=0.4\linewidth]{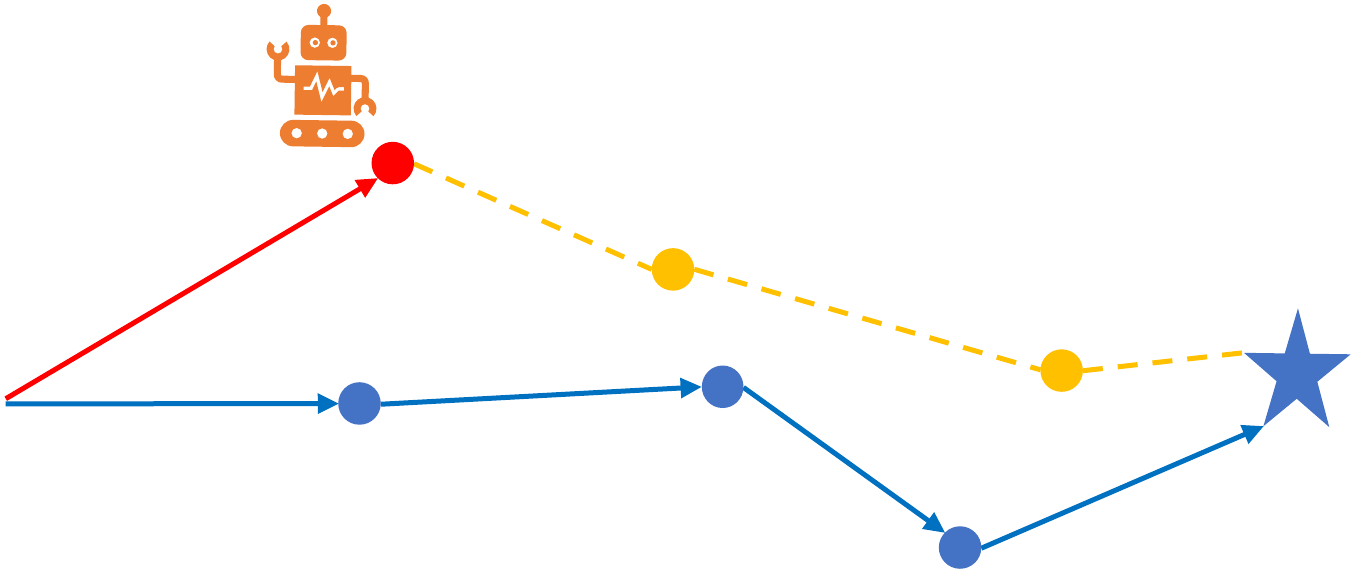}}%
	\hfill
	\subfigure[Distance to Nearest Point]{
		\label{nearest_pathloss}
		\includegraphics[width=0.4\linewidth]{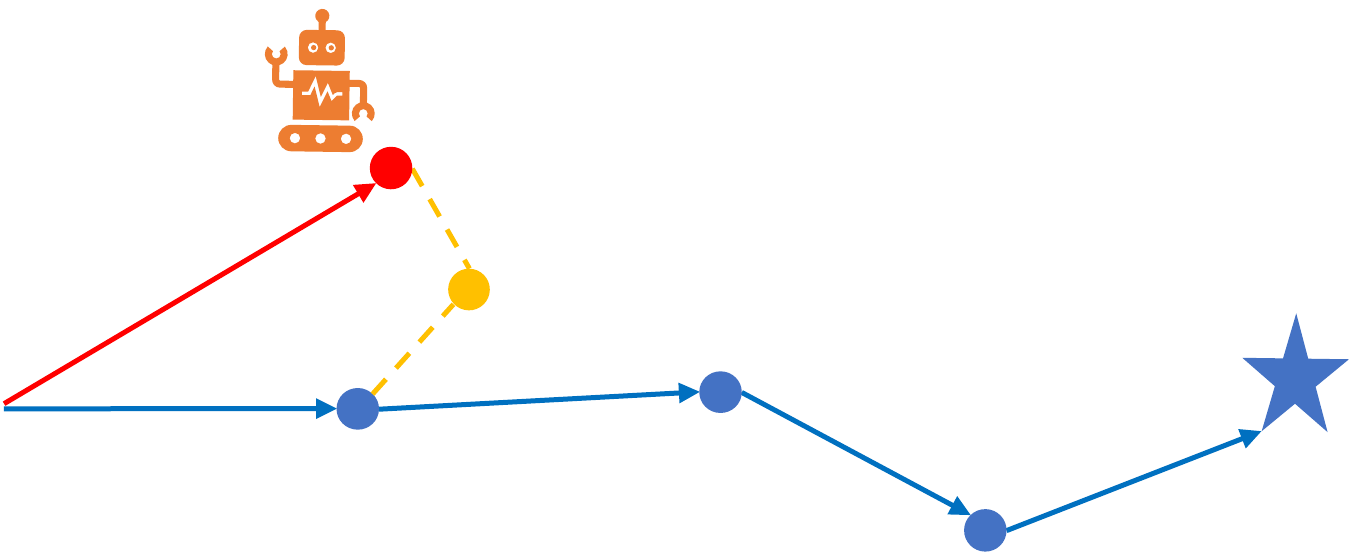}}%
	\hfill
	\caption{Illustration of distance to target (a) and distance to nearest point(b). The trajectory of the agent is in red and the ground-truth path is in blue. The star indicates the target viewpoint.  The brown dot line indicates the shortest trajectory length.}
	\label{rlloss}
\end{figure}

Following the training process of our baseline~\cite{envdrop},  the model is trained using both  imitation learning (IL) and reinforcement learning (RL) with original training data and augmented data. In addition to the losses in the baseline, we introduce a Nearest Point Loss (NPL) to encourage the agent to stay on ground-truth paths. 
NPL is based on the distance to the nearest viewpoint on the ground-truth path (see Figure~\ref{nearest_pathloss})
\begin{equation}
\mathcal{L}^{NP}=\sum\nolimits_{t}\textrm{log}(p_t(a_t))*D^{NP}_t,
\end{equation}
where $D^{NP}_t=\min_{\textrm{P}_i\in \textrm{Q}}d(\textrm{P}_t,\textrm{P}_i)$, $d(\textrm{P}_t,\textrm{P}_i)$ is the shortest trajectory distance between the current viewpoint $\textrm{P}_t$ and each viewpoint $\textrm{P}_i$ on the ground truth path, $\textrm{Q}$ is the set of viewpoints on the ground-truth path. If an agent stays on the ground-truth path, $D_t^{NP}$ would be zero; otherwise, the farther the larger. 

%

\section{Experiments}

\subsection{Dataset and Implementations}

\noindent\textbf{Dataset}
Experiments are conducted on the widely used indoor R2R ~\cite{r2r} and the R4R~\cite{r4r} VLN datasets.  R2R dataset consists of train/seen validation/unseen validation/test splits, involving 7,189 paths with 21,567 human annotated instructions with an average length of 29 words. In particular, these four splits contains 14,025/1,020/4,173/2,349 instructions, respectively. R4R aims to evaluate whether an agent is able to follow instructions by concatenating several trajectories and thus instructions of R2R (the shortest path is therefore no longer the ground-truth path) into long ones. Note that the R4R dataset only contains train/seen validation/unseen validation splits.

\noindent\textbf{Implementation Details}
Following the common practice, we use the pre-computed   ResNet~\cite{resnet} feature to represent percepted images. We adopt the same training strategy as our baseline~\cite{envdrop}: First, we train the agent using real data (i.e., the training split); Then, we train the agent using augmented data generated by the speaker and environment dropout techniques. 
We implement the proposed method with PyTorch, and all the experiments are carried out on NVIDIA GeForce RTX 2080 Ti. 

\subsection{Evaluation Metrics}
To evaluate the proposed method on R2R, we adopt four metrics from \cite{eval,r2r}: Success Rate (SR), Trajectory Length (TL), Oracle Success Rate (OSR), and Success rate weighted by Path Length (SPL). Among the four metrics, SPL is the main metric because it emphasizes the trade-off between SR and TL, which is also a recommended primary measure of navigation performance by the official VLN Challenge~\cite{eval}.  To measure the performance on R4R, we additionally adopt  the Coverage weighted by Length Score (CLS) metric as recommended by the builder of this dataset~\cite{r4r} and the Success weighted by normalized Dynamic Time Warping (SDTW) metric as recommended in~\cite{sdtw}.  These two metrics focus on navigation fidelity using the ground-truth path as reference.

\begin{table*}[!t]
	\resizebox{\textwidth}{!}{
		\begin{tabular}{ccc|cccccc|cccccc}
			\toprule
			\multicolumn{3}{c|}{\textbf{Model}} & \multicolumn{6}{c|}{\textbf{Val Seen}} & \multicolumn{6}{c}{\textbf{Val Unseen}} \\ 
			\hline
			Baseline& OAAM & NPL & SR$\uparrow$ & OSR$\uparrow$ & SDTW$\uparrow$ & SPL$\uparrow$ & TL$\downarrow$ & CLS$\uparrow$ & SR$\uparrow$ & OSR$\uparrow$ & SDTW$\uparrow$ & SPL$\uparrow$ & TL$\downarrow$ & CLS$\uparrow$ \\ \hline
			\cmark& & & 0.63 & 0.70 & 0.53 & 0.60 & 10.17 & 0.70 & 0.50 & 0.57 & 0.37 & 0.47 & 9.27 & 0.60 \\
			\cmark&\cmark &  & 0.65 & 0.72 & 0.53 & 0.62 & 10.21 & 0.71 & 0.51 & 0.59 & 0.37 & 0.47 & 10.05 & 0.60 \\
			\cmark&&\cmark  & 0.68 & 0.74 & 0.57 & 0.66 & 10.06 & 0.73 & 0.48 & 0.54 & 0.35 & 0.45 & 9.04 & 0.60 \\
			\cmark&\cmark&\cmark & 0.65 & 0.73 & 0.53 & 0.62 & 10.20 & 0.73 & 0.54 & 0.61 & 0.39 & 0.50 & 9.95 & 0.61 \\ 
			\bottomrule
		\end{tabular}
	}
	\caption{Ablation study of the proposed method on the R2R dataset (SPL is the main metric). The proposed object-and-action aware model (OAAM) and nearest point loss (NPL) mainly improve performance on the Unseen and Seen scenarios, respectively. Their combination leads to a better performance in unseen environments.}
	\label{tab:ablation_modules}
\end{table*}

\subsection{Ablation Study}
We conduct the ablation study to  find out the effectiveness of the proposed object-and-action aware model (OAAM) and the nearest point loss (NPL). The results are presented in Table~\ref{tab:ablation_modules}. 
The results show that: (I) Both OAAM and NPL improve the performance on the Val Seen split with about $2\%$ and $4\%$ increase in SPL, respectively. (II) The phenomenon on Val Unseen is a bit complicated. When OAAM or NPL works alone with the baseline, OAAM brings about $1\%$ improvement in SR and  NPL harms the SR about $2\%$. When OAAM and NPL work together, the performance is significantly improved about $4\%$ in SR and $3\%$ in SPL. This can be attributed to that NPL is able to help OAAM to find viewpoints on the ground-truth path and therefore improves the SR as well as shortens the trajectory length compared to the case that they work separately. Overall, both OAAM and NPL facilitate our method to achieve better performance, especially in the unseen scenario.

We further study the importance of the object-aware (OA) module and the action-aware (AA) module within the trained OAAM.  
As these two sub-modules contribute to the final navigation decision via a weighted sum~\eqref{weightedsum}, we test the performance of each module by setting its weight to 1 and the other weight to 0. The results are presented in Table~\ref{tab:ablation_submodules}. There is about $20\%$ performance drop in SR when OA or AA works alone. 
This indicates that both modules contribute to the final performance and the adaptive combination module  plays a crucial role (also see visualized samples in Figure~\ref{fig:example} and \ref{fig:example_2}).  Furthermore,  the AA module performs consistently better  than the object-aware module on ValSeen, ValUnSeen, and Test splits, in terms of SPL, SR, or OSR. For example, when AA module is removed, the SR is about $20\%$ lower than that when OA module is removed on the Unseen split. This indicates the AA module is more generalizable than the OA, which is roughly consistent with the claim that visual feature may hurt models in unseen environments~\cite{ryl}.  Fortunately, when OA and AA are adaptively combined, a much better performance is achieved ($10\%\sim28\%$ improvement in SPL on the Val Unseen split). Additionally, we calculate the mean of $w_t$ of all steps on ValSeen and ValUnseen. Results show that OA and AA are asigned 0.18 and 0.82 on average, respectively. This also reflects AA contributes more to the final results. Lastly, we make a try to conduct quantitative evaluation of these two modules based on NLTK tags, although it is not perfect. An attention is considered success if the top3 words attended by OA contain nouns (verbs for AA). OA and AA achieve accuracy of about 80\% and 75\% on average, respectively. This at some extent indicates improvement room for these modules.

\begin{table*}[!t]
	\resizebox{\textwidth}{!}{
		\begin{tabular}{l|cccccc|cccccc}
			\toprule
			\multirow{2}{*}{\textbf{Model}} & \multicolumn{6}{c|}{\textbf{Val Seen}} & \multicolumn{6}{c}{\textbf{Val Unseen}} \\ \cline{2-13} 
			& SR$\uparrow$ & OSR$\uparrow$ & SDTW$\uparrow$ & SPL$\uparrow$ & TL$\downarrow$ & CLS$\uparrow$ & SR$\uparrow$ & OSR$\uparrow$ & SDTW$\uparrow$ & SPL$\uparrow$ & TL$\downarrow$ & CLS$\uparrow$ \\ \hline
			Full Model & 0.65 & 0.73 & 0.53 & 0.62 & 10.20 & 0.73 & 0.54 & 0.61 & 0.39 & 0.50 & 9.95 & 0.61 \\ 
			\;\;w/o AA Module& 0.42 & 0.52 & 0.31 & 0.39 & 9.26 & 0.59 & 0.26 & 0.34 & 0.15 & 0.22 & 8.86 & 0.46 \\
			\;\;w/o OA Module & 0.45 & 0.53 & 0.32 & 0.42 & 10.63 & 0.56 & 0.44 & 0.52 & 0.30 & 0.40 & 10.46 & 0.54 \\ 
			\bottomrule
		\end{tabular}
	}
	\caption{Comparison of object-aware module (OA) and action-aware module (AA)  on the R2R dataset. `w/o' denotes to remove the module from the full model. The AA module contributes more than the OA module. }
	\label{tab:ablation_submodules}
\end{table*}

\subsection{Comparison to State-of-The-Art Navigators}
In this section, we compare our model against six state-of-the-art navigating methods, including FAST~\cite{fast}, RCM~\cite{rcm}, EnvDrop~\cite{envdrop}, Speaker-Follower (SF)~\cite{sf}, RegretAgent (RA)~\cite{regret}, and Self-Monitor (SM)~\cite{selfMonitor}. Results on the R2R and R4R datasets are presented in Table~\ref{tab:r2r} and \ref{tab:r4r}, respectively.

As shown in Table \ref{tab:r2r}, our model achieves the best performance on both the Val UnSeen and Test splits in terms of the main evaluation metric, SPL, which is up to $50\%$. 
As analysed in the ablation study, the performance of our model is the result of both the OAAM and the NPL loss. 
We also observe that on the Val Seen split, our model ranks third falling behind FAST and RA. This indicates that our model may not fit the training data as well as FAST and RA, and there are more information can be learned from the training data for our model. We leave this for the future exploration.

Table~\ref{tab:r4r} shows the result on the R4R dataset, where CLS is the recommended metric~\cite{r4r}.  On one side, the proposed method performs consistently better than our baseline, namely Envdrop, with up to $6\%$ improvement in CLS on the unseen split. On the other side, the proposed method achieves the best performance with a $0.40$ CLS score in unseen environments, which is $3\%$ higher than the second best and thus set the new SoTA.  Similar to the phenomenon on the R2R dataset, we also observe that the proposed model ranks top three in the seen environments, which indicates more information could be learned.

\begin{table*}[!t]
	\resizebox{\textwidth}{!}{
		\begin{tabular}{@{}l|cccc|cccc|cccc@{}}
			\toprule
			\multirow{2}{*}{\textbf{Model}} & \multicolumn{4}{c|}{\textbf{Val Seen}} & \multicolumn{4}{c|}{\textbf{Val Unseen}} & \multicolumn{4}{c}{\textbf{Test}} \\ \cline{2-13}
			& SR$\uparrow$ & OSR$\uparrow$ & SPL$\uparrow$ & TL$\downarrow$ & SR$\uparrow$ & OSR$\uparrow$ & SPL$\uparrow$ & TL$\downarrow$ & SR$\uparrow$ & OSR$\uparrow$ & SPL$\uparrow$ & TL$\downarrow$ \\ \hline
			SF~\cite{sf} & 0.67 & 0.74  & 0.61 & 11.73 & 0.35 & 0.45 & 0.28 & 14.56  & 0.35 & 0.44 & 0.28 & 14.82 \\
			SM~\cite{selfMonitor} & 0.70 & 0.80  & 0.61 & 12.90  & 0.43 & 0.54 & 0.29 & 17.09 & 0.48 & 0.59 & 0.35 & 18.04 \\
			RA~\cite{regret} & 0.69 & 0.77  & \color{green}0.63 & -  & 0.50 & 0.59 & 0.41 & -  & 0.48 & 0.56 & 0.40 & 13.69 \\
			RCM~\cite{rcm} & 0.67 & 0.75  & - & 10.65 & 0.43 & 0.50 & - & 11.46  & 0.43 & 0.50 & 0.38 & 11.97 \\
			FAST~\cite{fast} & 0.70 & 0.80  & \color{red}0.64 & 12.34 & 0.56 & 0.67 & \color{blue}0.44 & 20.45 & 0.54 & 0.64 & \color{blue}0.41 & 22.08 \\
			EnvDrop~\cite{envdrop} & 0.63 & 0.70  & 0.60 & 10.17 & 0.50 & 0.57 & \color{green}0.47 & 9.27  & 0.50 & 0.57 & \color{green}0.47 & 9.70 \\ \hline
			\textbf{Ours} & 0.65 & 0.73 & \color{blue}0.62 & 10.20 & 0.54 & 0.61 & \color{red}0.50 & 9.95 & 0.53 & 0.61 & \color{red}0.50 & 10.40 \\ 
			\bottomrule
		\end{tabular}
	}  
	\caption{Comparison against several state-of-the-art VLN methods on the R2R dataset. SPL is the main metric. The best three results are highlighted in red, green, and blue fonts, respectively.}
	\label{tab:r2r}
\end{table*}

\begin{table*}[!t]
	\centering
	\resizebox{\textwidth}{!}{%
		\begin{tabular}{@{}l|ccccc|ccccc@{}}
			\toprule
			\multirow{2}{*}{\textbf{Model}} & \multicolumn{5}{c|}{\textbf{Val Seen}} & \multicolumn{5}{c}{\textbf{Val Unseen}} \\ \cline{2-11} 
			& SR$\uparrow$  & SDTW$\uparrow$ & SPL$\uparrow$ & TL$\downarrow$ & CLS$\uparrow$ & SR$\uparrow$ & SDTW$\uparrow$ & SPL$\uparrow$ & TL$\downarrow$ & CLS$\uparrow$ \\ \hline
			SF~\cite{sf} & 0.52 & - & 0.37 & 15.40 & 0.46 & 0.24 & - & 0.12 & 19.90 & 0.30 \\
			RCM goal-oriented~\cite{r4r} & 0.56 & - & 0.32 & 24.50 & 0.40 & 0.29 & - & 0.10 & 32.50 & 0.20 \\
			RCM fidelity-oriented~\cite{r4r} & 0.53 & - & 0.31 & 18.80 & \color{green}0.55 & 0.26 & - & 0.08 & 28.50 &\color{blue}0.35 \\
			PTA high-level \cite{pta} & 0.58 &  0.41 & 0.39 & 16.50 & \color{red}0.60 & 0.24 &  0.10 & 0.10 & 17.70 & \color{green}0.37 \\ 
			EnvDrop \cite{envdrop} & 0.52 & 0.27& 0.41 & 19.85 & 0.53 & 0.29 &0.09&0.18&26.97 &0.34\\
			\hline
			\textbf{Ours} & 0.56 &  0.32 & 0.49 & 11.75 & \color{blue}0.54 & 0.31 &  0.11 & 0.23 & 13.80 & \color{red}0.40 \\
			\bottomrule
		\end{tabular}%
	}
	\caption{Results on the R4R dataset. CLS is the main metric. The best three results are highlighted in red, green, and blue fonts, respectively.}
	\label{tab:r4r}
\end{table*}

\subsection{Qualitative Results}
\begin{figure*}[!b]
	\centering
	\includegraphics[width=\textwidth]{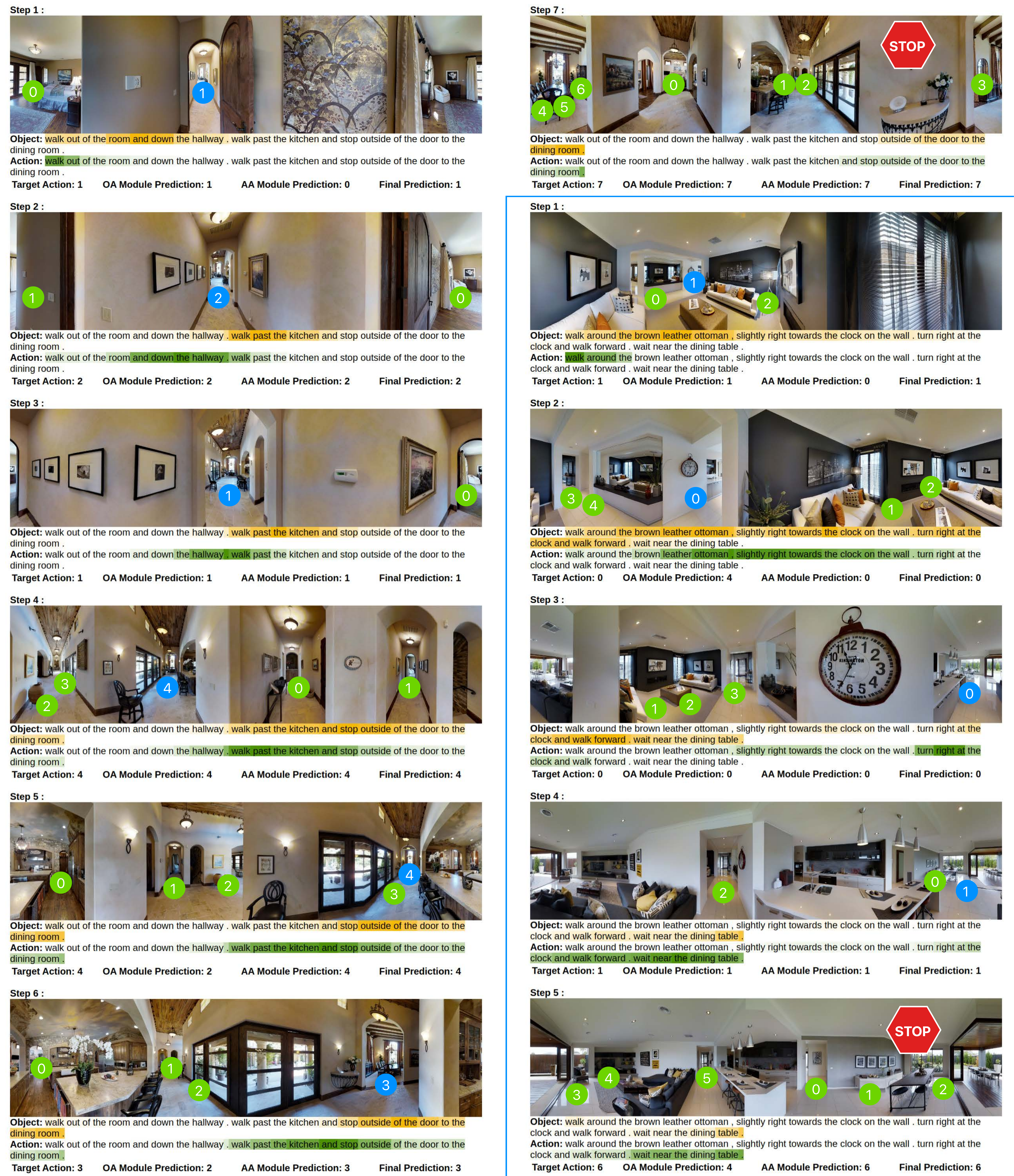}
	\caption{Navigation samples of the proposed method in two unseen environments.  The dark or light color on the instruction denotes the relative attention value (the darker the larger). {The numbered circles} denote navigable viewpoints, and the blue ones denote the final decision. }
	\label{fig:example}
\end{figure*}

\begin{figure*}[!t]
	\centering
	\includegraphics[width=\textwidth]{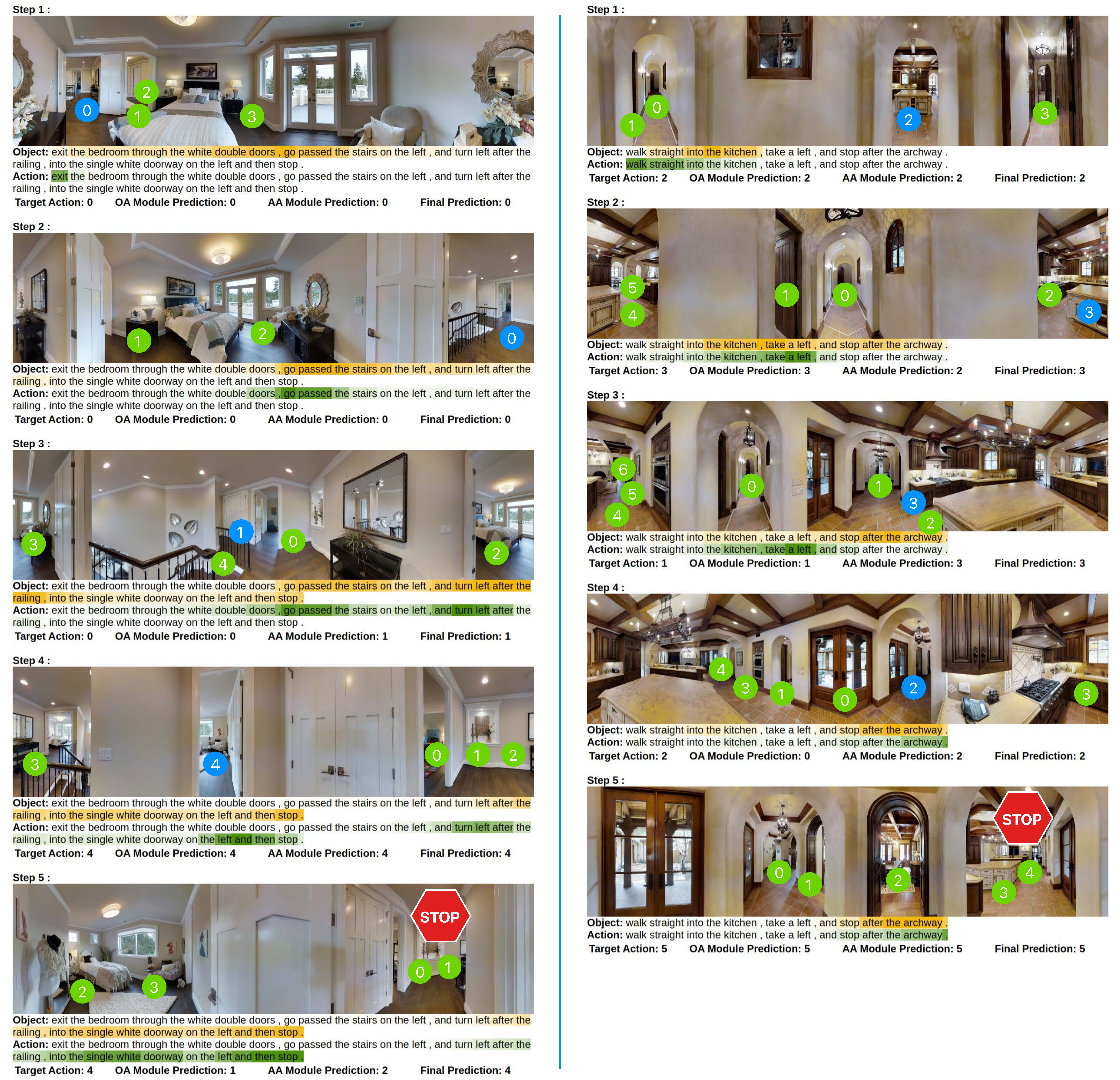}
	\caption{Two more navigation samples of the proposed method in unseen environments. On the left panel, at step 3, our model predicts an incorrect action but gets to the goal room in end. It is worth noting that at step 5, both OA and AA modules give wrong prediction, but after combination our model predicts the correct action. On the right panel, the final decision at each step is correct, but we could observe that the OA or AA module not always gives correct prediction, such as at step 2 and step 3. These indicate that all the OA and AA modules as well as the adaptive combination module are crucial to the final success.
	}
	\label{fig:example_2}
\end{figure*}

In this section, we visualize several navigation process at each step in Figure~\ref{fig:example} and Figure~\ref{fig:example_2}, including key information, such as attention distribution and navigation decision. 

For the trajectories in Figure~\ref{fig:example}, most of the attention predicted by the OA and AA modules are  able to correctly reflect the object or action that should be considered to make correct decisions. At some midway steps, such as step 1 and 5 on the left panel of Figure~\ref{fig:example},  only one module gives the correct prediction but the final prediction is correct after combination. We also show examples that some midway decisions are incorrect but finally get to the goal room in Figure~\ref{fig:example_2}, such as step 3 in the left panel. 

\section{Conclusion}

In this paper, we propose an object-and-action aware model for robust visual-and-language navigation. Object and action information in instructions are separately highlighted, and are then matched to visual perceptions and orientation embeddings, respectively. To encourage the robot agent to stay on the path, we additional propose a path loss based on the distance to nearest ground-truth viewpoint. Extensive experimental results demonstrate the superiority of our model compared against several state-of-the-art navigating methods, especially on the unseen test scenarios.

\clearpage
%
%
\bibliographystyle{splncs04}
\bibliography{egbib}
\end{document}